\documentclass{article}
\usepackage{ijcai15}
\usepackage{times}
\usepackage{url}
\usepackage{latexsym}
\usepackage{url}

\renewcommand{\c}{\ensuremath{\mathbf{c}}}

\newcommand{\x}{\ensuremath{\mathbf{x}}}
\newcommand{\y}{\ensuremath{\mathbf{y}}}

\newcommand{\calC}{\ensuremath{\mathcal{C}}}
\newcommand{\calD}{\ensuremath{\mathcal{D}}}

\newcommand{\calG}{\ensuremath{\mathcal{G}}}

\newcommand{\calL}{\ensuremath{\mathcal{L}}}

\newcommand{\calW}{\ensuremath{\mathcal{W}}}
\newcommand{\calX}{\ensuremath{\mathcal{X}}}
\newcommand{\calY}{\ensuremath{\mathcal{Y}}}

\usepackage{graphicx}
\usepackage{amsfonts, amssymb, amsmath}
\usepackage{algorithmic, amsthm}

\newtheorem{prop}{Proposition}[section]
\usepackage[ruled,vlined]{algorithm2e}
\usepackage{mathrsfs}

\title{Syntax-based Deep Matching of Short Texts\thanks{this work is done when the first author worked as intern at Noah's Ark Lab, Huawei Technologies.}}

\author{Mingxuan Wang$^1$ \ Zhengdong Lu$^2$ \ 
	Hang Li$^2$ \ Qun Liu$^{3,1}$ \\
	$^1$Institute of Computing Technology, Chinese Academy of Sciences\\
	$^2$Noah's Ark Lab, Huawei Technologies\\
	$^3$Centre for Next Generation Localisation, Dublin City University\\
}

\date{}

\begin{document}
\maketitle
\begin{abstract}
Many tasks in natural language processing, ranging from machine translation to question answering, can be reduced to the problem of matching two sentences or more generally two short texts. We propose a new approach to the problem, called Deep Match Tree (\textsc{DeepMatch}$_{tree}$), under a general setting. The approach consists of two components, 1) a mining algorithm to discover patterns for matching two short-texts, defined in the product space of dependency trees, and 2) a deep neural network for matching short texts using the mined patterns, as well as a learning algorithm to build the network having a sparse structure. We test our algorithm on the problem of matching a tweet and a response in social media, a hard matching problem proposed in \cite{emnlpmatch}, and show that \textsc{DeepMatch}$_{tree}$ can outperform a number of competitor models including one without using dependency trees and one based on word-embedding, all with large margins.
\end{abstract}


\section{Introduction} 
Matching is of central importance to natural language processing. In fact, many problems in natural language processing can be formalized as matching between two short-texts, with different matching relations in different applications. For example, in paraphrase identification the relation is synonymy, and in information retrieval it is relevance. In the meantime matching is also a challenging problem, since it requires modeling of the two short-texts as well as their relation. In machine translation, for example, the model needs to determine whether a sentence in the source language has the same meaning as a sentence in the target language. In dialogue, the model needs to judge whether a message is an appropriate response to a given utterance.

Deep neural network can model non-linear and hierarchical relations~\cite{DeepAI}, and thus is well suited for short-text matching in natural language processing. The very limited work in that thread, makes use of word embedding as the building blocks of matching model. Although embedding-based methods have been proven effective on tasks like question answering~\cite{nipsmatch}, paraphrase identification~\cite{SocherPool}, and even short text conversation \cite{nipsmatch,NIPS2014_5550}, they are not enough good at handling the subtlety of general short-text matching. Short-texts often represent rich content, their relations are also complicated, and more sophisticated structures are required for comparing the two short-texts. For example, when judging the appropriateness of response ``\texttt{\small You should rest more.}" to utterance ``\texttt{\small I have to work during the weekend!}", we have to consider the semantic correspondence between ``work over the weekend" and ``need to rest more", which is hard to be captured by an embedding-based model.

We study the problem of short-text matching in a general setting. Our method, named \emph{Deep Match Tree} (\textsc{DeepMatch}$_{tree}$), consists of two sequentially connected components: 1) a mining algorithm to discover rich yet subtle patterns, defined in the product space of dependency trees, from a large corpus of paired short-texts, and 2) a learning algorithm to construct a deep neural network (DNN) for making a matching decision on the two short-texts, on the basis of the mined patterns. The DNN model is specifically trained based on contrastive sampling of negative examples.

Without loss of generality, we focus on the task of matching a response to a given tweet on Weibo, a popular Chinese microblog service, for which a large amount of data is available. This is a hard problem, requiring consideration of complicated correspondence between the structures of two texts. Our experimental results show that \textsc{DeepMatch}$_{tree}$ is superior to existing methods on the problem.

Our main contributions are: 1) proposal of an algorithm for mining dependency tree matching patterns on large scale, 2) proposal of an algorithm for learning a deep matching model for using mined matching patterns, and 3) empirical validation of the efficacy and efficiency of the proposed method using large scale real datasets. 

\begin{figure*}[t!]
	\begin{center}
		\begin{tabular}[c]{cc}
			\includegraphics[width=.75\textwidth]{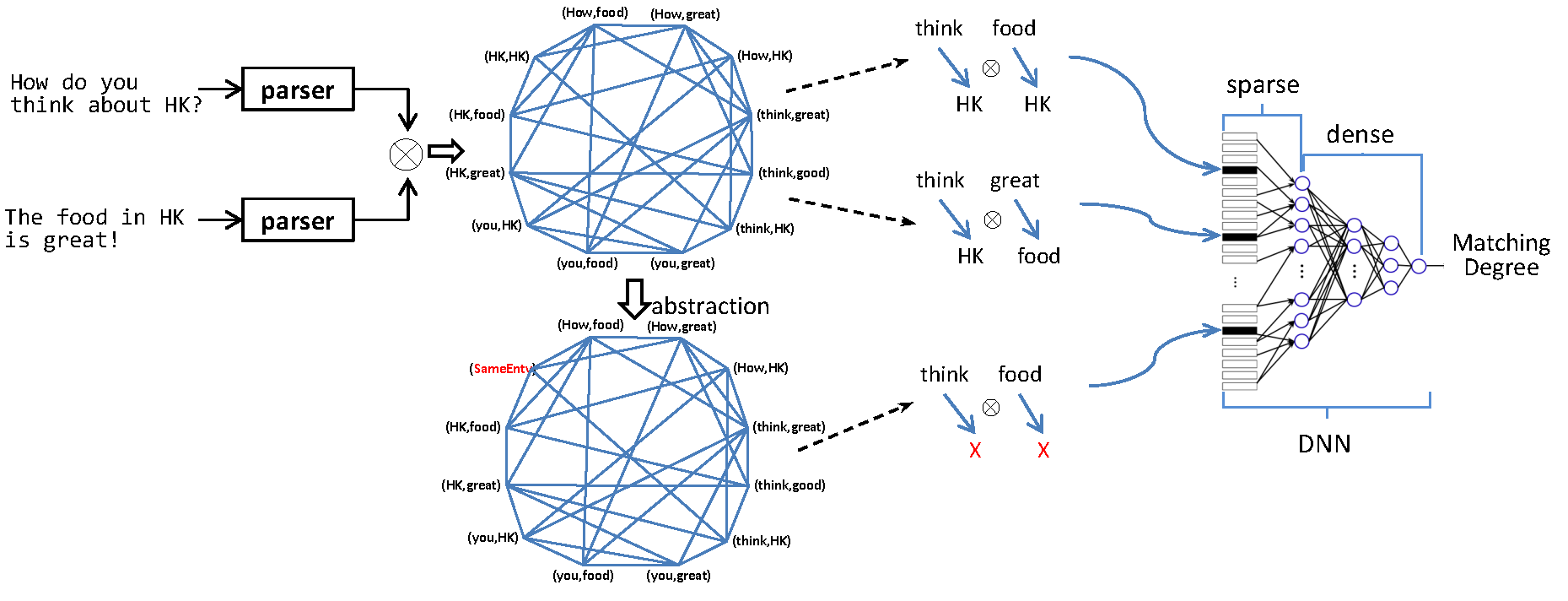}
		\end{tabular} 
		\caption{The overall architecture for \textsc{DeepMatch}$_{tree}$.}
		\label{f:OverallDiagram}
	\end{center}
\end{figure*} 

\section{Direct Product of Graphs (PoG)} \label{s:PoG} 
We first propose representing the matching of a pair of sentences (in general short-texts) with the direct product between the dependency trees of them, and then propose treating subgraphs of this product graph as matching patterns. 

\subsection{Dependency Tree}

\begin{figure}[h!]
\begin{center}
      \includegraphics[width=0.35\textwidth]{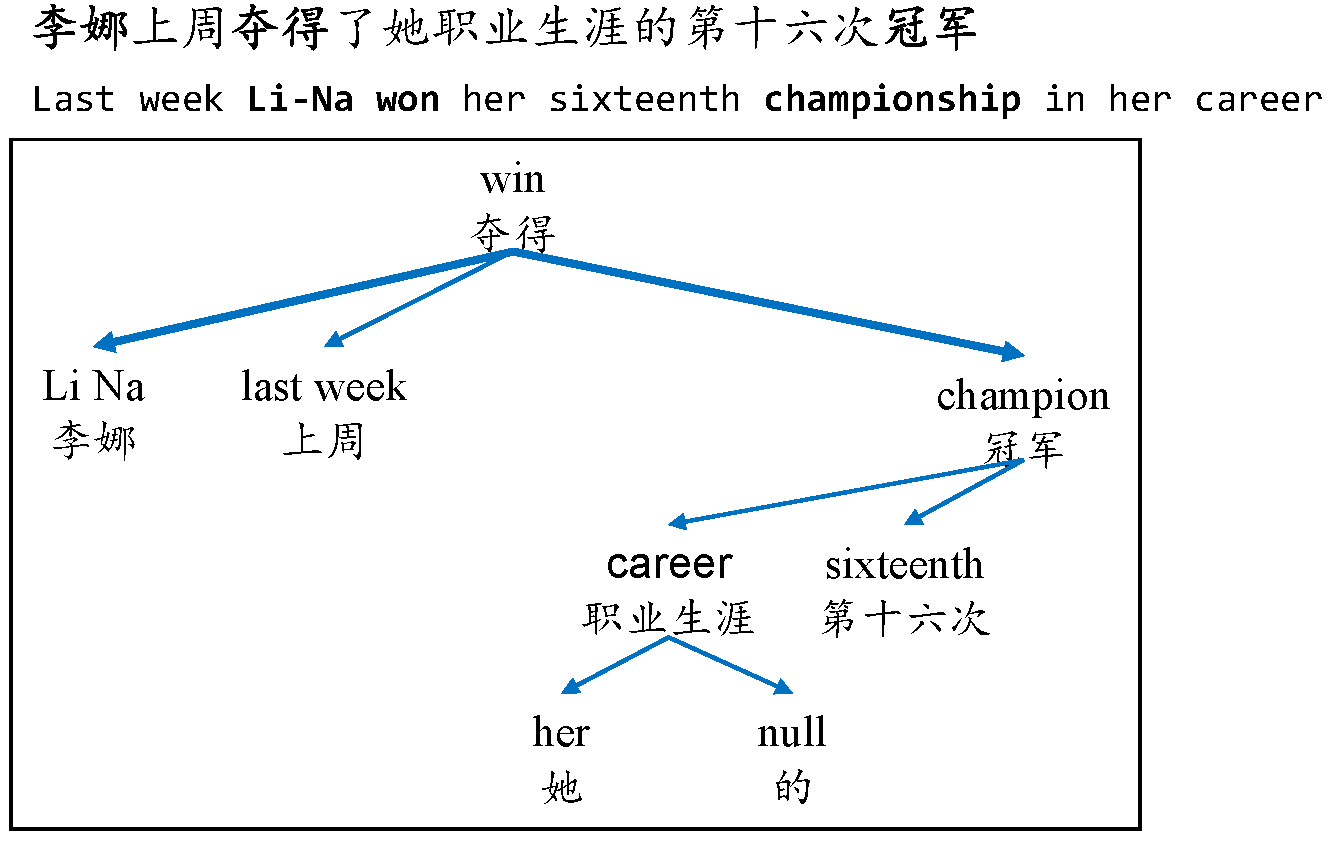}
    \caption{Example of dependency tree, where the main structure of the sentence is represented as the sub-tree in thick edges. The tweet is in Chinese (literal English translation).    }
    \label{f:dtree}
  \end{center}
\end{figure}

We  represent a sentence with its dependency tree. We choose to do so because a dependency tree tends to expose the ``skeleton" of the sentence, revealing both short-distance and long-distance grammatical relations between words~\cite{depconpression}.
For example the dependency tree in Fig.\ref{f:dtree} contains structures like \{$\texttt{\small Li Na}\hspace{-5pt}\leftarrow \hspace{-5pt}\texttt{\small win}\hspace{-5pt}\rightarrow\hspace{-3pt}\texttt{\small championship} $\} represented as a sub-tree,where the words (boldface) are not necessarily adjacent to each other in the sentence. 


\subsection{Direct Product of Graphs} 
The direct product of graphs (PoG) $\calG_X = \{V_X, E_X\}$ and $\calG_Y = \{V_Y, E_Y\}$, is a graph $\calG_{X\times Y}$ \cite{VishproG}, with vertices $V_{X \times Y}$ and edges $E_{X \times Y}$
{\small
	\begin{eqnarray*}
		V_{X \times Y}&=& \{(v^X_i, v^Y_{i'}), \; v^X_i \in V_X, \,v^Y_{i'} \in V_Y\}\\
		E_{X\times Y} &=& 
		\{((v^X_i, v^Y_{i'})  (v^X_j, v^Y_{j'})), \; (v^X_i, v^X_j)
		\\ & & \in E_X  \wedge (v^Y_{i'}, v^Y_{j'}) \in E_Y\}
	\end{eqnarray*}}

\begin{figure}[t!]
\begin{center}
\begin{tabular}[l]{l}
       \includegraphics[width=0.45\textwidth]{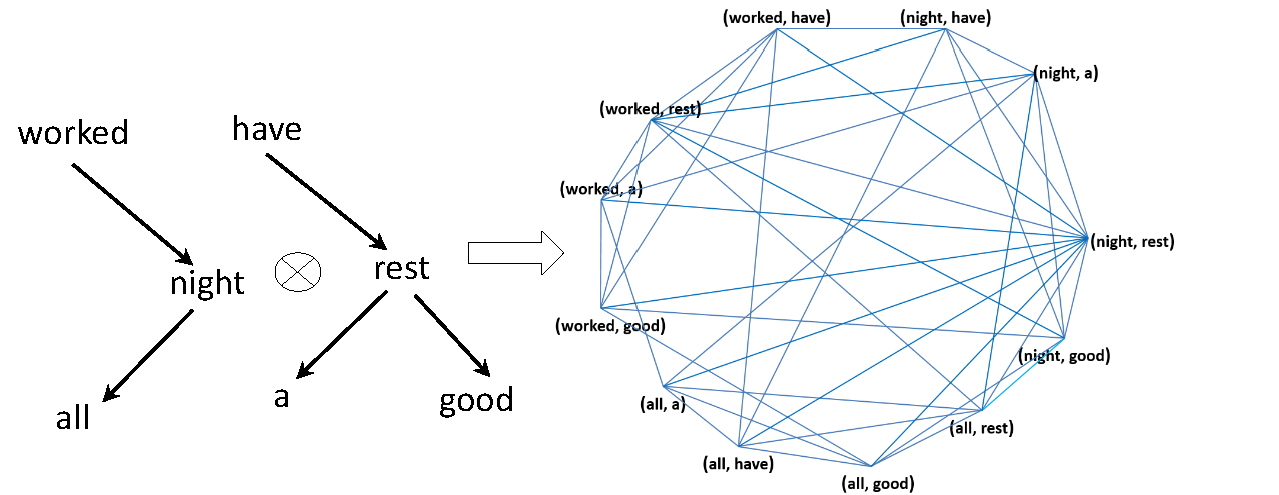}\\
\end{tabular}
\caption{The direct product of two dependency trees.}
 \label{f:arc}
\end{center} 
\end{figure}

\noindent Given two sentences $S_X$ and $S_Y$, their interaction relation is represented by the direct product of their dependency trees $\calG_{X\times Y}$.  For example, two sentences \texttt{\small Worked all night} and \texttt{\small Have a good rest} (with their dependency trees are given by the left panel of Fig.\ref{f:arc}), and the direct product of the trees is given by the right panel of Fig.\ref{f:arc}. Note that $\calG_{X\times Y}$  is in general a graph even though $\calG_X$ and $\calG_Y$ are trees.

$\calG_{X\times Y}$ directly describes the interaction relation between sentences $S_X$ and $S_Y$, hosting a rather rich set of structures, both lexical and syntactic, that contribute to the overall matching between the two sentences. Next we make further abstraction of the representation.

\subsection{Abstraction} \label{s:abstraction} 
We consider two types of abstraction for vertices in $\mathcal{G}_{X \times Y}$
\begin{itemize}
  \item {\bf Same Entity:} We replace the vertex $(ne_i, ne_i)$ in $\calG_{X \times Y}$ representing the same entity with a general vertex \textsc{ SameEntity}. For example for the sentences (\texttt{\small How is the wether in Paris?}) (\texttt{\small Haven't seen such a sunny day in Paris for a while!}), the vertex (\texttt{\small Paris}, \texttt{\small Paris}) after the abstraction will be treated as the same vertex as (\texttt{\small Boston}, \texttt{\small Boston}) after the same type of abstraction. Graph with this type of abstraction is named $\tilde{\calG}^{e}_{X \times Y}$.

  \item {\bf Similar Word:} We conduct clustering of words based on their word2vectors \cite{word2vec} using the K-means algorithm. For a vertex $(w_i, w_j)$ in the product graph, if $w_i$ and $w_j$ belong to the same word cluster $\calC_k$, then the vertex will be replaced with a new vertex $\textsc{SimWord}_k$. Graph with this type of abstraction is named $\tilde{\calG}^{s}_{X \times Y}$.
\end{itemize} 
Both types of abstraction will enhance the generalization ability of  matching pattern mining described next. 

\subsection{Sub-graphs of PoG as Matching Patterns} 
With a little abuse of notation, we use $\bar{\mathcal{G}}_{X \times Y} = \{\mathcal{G}_{X \times Y}, \tilde{\mathcal{G}}^e_{X \times Y}, \tilde{\mathcal{G}}^s_{X \times Y}\}$ to denote the PoG for sentence pair $(S_X, S_Y)$ as well as its variants after two types of abstraction. For a sentence pair $(S_X, S_Y)$, any sub-graph in the corresponding $\bar{\mathcal{G}}_{X \times Y}$ describes part of the interaction between the two sentences and therefore can contribute to the matching between the two. For instance, $\texttt{\small (weather, sunny)} \longleftrightarrow \textsc{\small SameEntity}$ is a sub-graph describing the matching between two sentences in a conversation about weather (see the example two paragraph ago). In general,  $\bar{\mathcal{G}}_{X \times Y}$  contains all the meaningful matching patterns for the task.

\section{Mining of Matching Patterns} \label{s:graphmine} 
It is the responsibility of a mining algorithm to discover those sub-graphs of $\{\bar{\mathcal{G}}_{X \times Y}\}$ that can work as matching patterns to discriminate matched sentence pairs from mismatched ones, measured in terms of discriminative ability (cf.,\cite{graphmining}). Discriminative roughly means it gives some evidence on matching, i.e., it appears in matched pairs more frequently than unmatched pairs. An efficient mining algorithm is vital to the success of this method, when the number of instances is of the order of $10^6$ and the number of mined patterns is of the order of $10^7$.

\subsection{Speeding-up the Mining Process} 
Fortunately, we can leverage the following fact with respect to sub-graphs in the PoG $\calG_{X\times Y}$ (without abstraction).
\begin{prop}
Any connected {sub-graph} $\calG^{s}_{X\times Y}$  in $\calG_{X\times Y}$  can uniquely determine a
minimal sub-tree in $\calG_{X}$ and a
minimal sub-tree $\calG_{Y}$,  whose direct product can cover the $\calG^{s}_{X\times Y}$.
\end{prop}
As it implies, the mining of sub-graphs in PoG of trees can be reduced to jointly selecting the sub-trees on two sides. This can not only greatly speed up the mining process, but also avoid finding patterns with duplicate functionality for matching. In the remainder of the paper, we will use $\texttt{\small sub-tree}_1 \otimes \texttt{\small sub-tree}_2$ to denote a tree-pair (separated by $\otimes$) mined from the PoG. This however does not apply to the more general case of $\bar{\calG}_{X\times Y}$ when some vertices are replaced with non-factorable variants, like \textsc{\small SimWord$_{123}$}, for which we have to introduce some new tricks.



\subsection{Mining without Abstraction} 
The algorithm for mining without abstraction, sketched in Algorithm 1, is to recursively grow the mined sub-graphs while maintaining its discriminative ability. It starts with the simplest pattern (1,1), standing for one-word tree on both $X$ side (tweet) and $Y$ side (response), and grows the mined trees recursively. In each growing step (\texttt{\small LeftExtend()} and \texttt{\small RightExtend()}), the size of sub-trees is increased by one on either $X$ side or the $Y$ side, followed by a filtering step to remove the found pairs with discriminative ability less than a threshold. The growing step is efficient, since we can limit the search for patterns of ($m, n+1$) from the candidates formed by merging patterns of ($m,n$). In practice the time for looking-up each sub-tree pair is almost constant with the help of Hashmap. The following table gives some examples of the matching patterns discovered by Algorithm 1.

\begin{algorithm} \small 
	\SetKwFunction{dis}{DiscriminativeFilter}
	\KwIn{
		$\mathcal{T}$: tree pairs for original (tweet, response), MaxSize
	}
	\KwOut{Set of mined features $\mathcal{F}$\;}
	\caption{Discriminative Mining of Parse Trees for Parallel Texts}
	\SetKw{Initinalize}{Initinalize}
	\SetKwFunction{ENQUEUE}{ENQUEUE}
	\SetKwFunction{DEQUEUE}{DEQUEUE}
	\SetKwFunction{LeftExtend}{LeftExtend}
	\SetKwFunction{RightExtend}{RightExtend}
	\SetKwFunction{Empty}{Empty}
	\Initinalize $\;$
	$\mathcal{F} \leftarrow \emptyset;\; \mathcal{M} \leftarrow \emptyset;  \;$
	$Q\leftarrow [\,];\;\; $\\
	\ENQUEUE(Q, (1,1))\;
	\ForEach{ node set $x,y$ in $\mathcal{X}, \mathcal{Y}$}{
		Append each element of $x \otimes y$ to $\mathcal{F}_{1,1} $\;
		Append $x \otimes y$ to $\mathcal{M}_{1,1}$\;
	}
	$\mathcal{F}_{1,1} \leftarrow$ \dis{$\mathcal{F}_{1,1}$}\;
	\While{Q $\neq [\,]$}{
		$m,n$ $\leftarrow$ \DEQUEUE(Q)\;
		\If{$m+1<$  MaxSize $\wedge$  $(m+1,n)$ has not been processed}{
			[$\mathcal{M}_{m+1,n}, \mathcal{F}_{m+1,n}] \leftarrow $ \LeftExtend{$\mathcal{M}_{m,n}$}\;
			\ENQUEUE(Q, $(m+1,n)$)\;
			$\mathcal{F} \leftarrow \mathcal{F}\cup\mathcal{F}_{m+1,n}$\;
		}
		\If{$n+1<$ MaxSize $\wedge$  $(m,n+1)$ has not been processed}{
			[$\mathcal{M}_{m,n+1}, \mathcal{F}_{m,n+1}]  \leftarrow$ \RightExtend{$\mathcal{M}_{m,n}$}\;
			\ENQUEUE(Q, $(m,n+1)$)\;
			$\mathcal{F} \leftarrow \mathcal{F}\cup \mathcal{F}_{m,n+1}$\;
		}
	}
\end{algorithm}

\begin{figure*}[t!]
\begin{center}
      \includegraphics[width=0.8\textwidth]{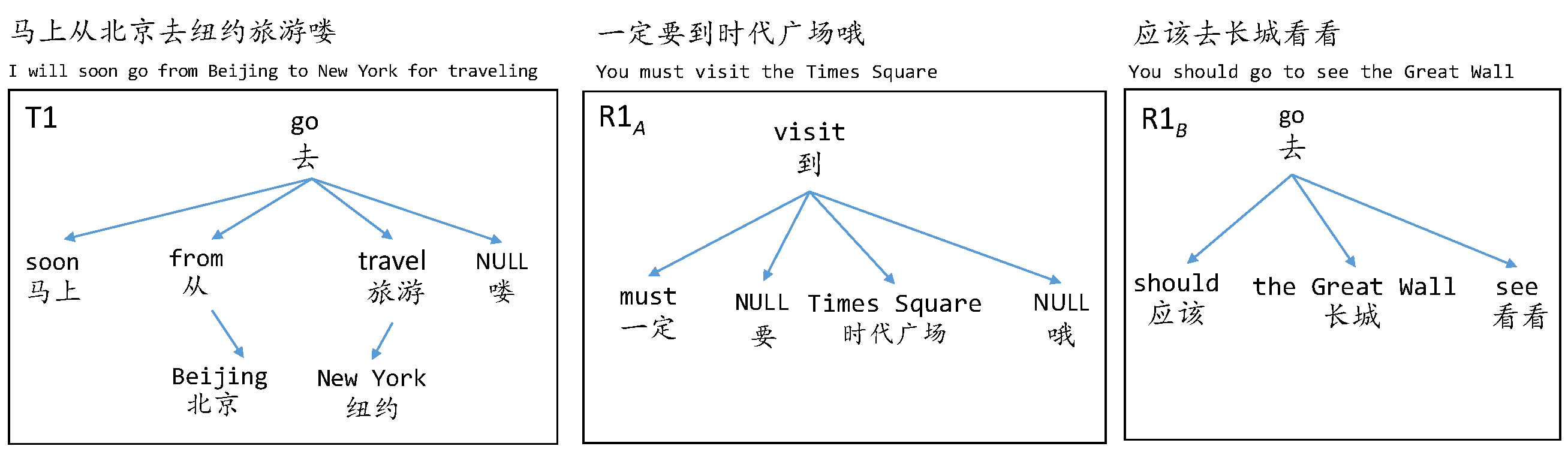}\\
    \caption{Illustration for dependency trees for tree short sentences.}
    \label{f:3TreeExamples}
  \end{center} 
  \end{figure*}

\begin{center}

\begin{tabular}{|l|}
  \hline
  Patterns without abstraction \\
  \hline
  $\texttt{\small exam}\; \otimes \; \texttt{\small score} $ \\
   \hline
  $\texttt{\small Information theory}\; \otimes \; \texttt{\small Shannon} $ \\
   \hline
  $\texttt{\small thank}\hspace{-3pt} \rightarrow \hspace{-3pt}\texttt{\small present}  \; \otimes \; \texttt{\small happy} \hspace{-3pt}\rightarrow \hspace{-3pt} \texttt{\small birthday}$ \\
  \hline
   $\texttt{\small win}\hspace{-3pt} \rightarrow \hspace{-3pt}\texttt{\small game}  \; \otimes \; \texttt{\small trying} \hspace{-3pt}\rightarrow \hspace{-3pt} \texttt{\small keep}$ \\
  \hline
$\texttt{\small out-of-control}\hspace{-3pt} \rightarrow \hspace{-3pt}\texttt{\small prices} \; \otimes \; \texttt{\small regulation} $ \\
\hline
$\texttt{\small work} \hspace{-3pt} \rightarrow \hspace{-3pt}\texttt{\small weekend} \; \otimes \; \texttt{\small rest} $ \\
\hline
  \end{tabular}
\end{center}

\subsection{Mining with Abstraction} 
The algorithm for mining with abstraction is a variant of Algorithm 1. Taking the SameEntity abstraction as example, we first replace each named entity, e.g., \texttt{\small Li Na} (found via a named entity resolution program) with a vertex having the same ID (say, \textsf{\small NamedEntity239}). The growing step is the same as in Algorithm 1, except that when counting the support (number of instances containing it) of a pattern, it replaces the same entity appearing on both sides with a wildcard, and therefore groups many patterns as the same one. For example, the instances for the following two patterns
\begin{eqnarray*}
	\texttt{\small Li Na}\leftarrow \texttt{\small win} \hspace{-3pt} &\otimes& \hspace{-3pt}  \texttt{\small Li Na}\leftarrow \texttt{\small congratulations}\\
	\texttt{\small Nadal}\leftarrow \texttt{\small win} \hspace{-3pt} &\otimes&  \hspace{-3pt} \texttt{\small Nadal}\leftarrow  \texttt{\small congratulations} 
\end{eqnarray*}
will be counted together for the pattern 
\begin{eqnarray*}
	x\leftarrow \texttt{\small win} &\otimes&   x\leftarrow  \texttt{\small congratulations} 
\end{eqnarray*} 
where $x$ stands for the wildcard. The mining with SimilarWord abstraction is similar, only slightly more complicated on deciding when two words can be merged.

The following table gives some examples of the matching patterns discovered by our algorithm on the graph with abstraction. Here $(x,x')$ stand for wildcards considered similar enough by the algorithm.
\begin{center}
\begin{tabular}{|l|}
  \hline
  Patterns with abstraction \\
  \hline
 $\texttt{\small hope}\hspace{-3pt} \rightarrow \hspace{-3pt}\texttt{\small win} \hspace{-3pt}\rightarrow \hspace{-3pt} x \; \otimes \; \texttt{\small support} \hspace{-3pt}\rightarrow \hspace{-3pt}x$ \\
  \hline
$\texttt{\small how about}\hspace{-3pt} \rightarrow  \hspace{-3pt} x \; \otimes \; \texttt{\small like} \hspace{-3pt}\rightarrow \hspace{-3pt}x$ \\
  \hline
$\texttt{\small gift} \hspace{-3pt} \rightarrow  \hspace{-3pt} x \; \otimes \; \texttt{\small happy} \hspace{-3pt}\rightarrow \hspace{-3pt}x$ \\
\hline
$\texttt{\small recommend} \hspace{-3pt} \rightarrow  \hspace{-3pt} x \; \otimes \; x \hspace{-3pt}\rightarrow \hspace{-3pt} \texttt{\small nice} $ \\
  \hline
  $\texttt{\small pretty good} \hspace{-3pt} \rightarrow  \hspace{-3pt} x \; \otimes \; \texttt{\small fine} \hspace{-3pt}\rightarrow \hspace{-3pt} \texttt{\small also} \hspace{-3pt}\rightarrow \hspace{-3pt} x'$ \\
\hline
\end{tabular}
\end{center}
\subsection{Advantage of Tree Pattern Mining}
It is important to note that dependency tree matching patterns can provide better correspondence between the two sentences than word co-occurrences  on two sides (an idea first explored in \cite{nipsmatch}).  To illustrate the superiority of using dependency tree matching patterns, suppose that for the tweet \textsf{\small T1} in Fig.\ref{f:3TreeExamples} we want to pick a more appropriate one from the two responses  (\textsf{\small R1$_A$} and \textsf{\small R1$_B$}). The word-based model tends to assign a high matching score to pair (\textsf{\small T1}, \textsf{\small R1$_B$}), due to the pattern $\{\small \texttt{Beijing, travel}\}\otimes\{\small \texttt{Great Wall}\}$, which is however spurious since \textsf{\small T1} is about traveling to New York while the word \texttt{\small Beijing} is a distractor. On the other hand, the tree-based model relies more on patterns like follows 
\begin{eqnarray*}
  \texttt{\small travel}\rightarrow \texttt{\small New-York} &\otimes& \texttt{\small Times Square}  \\
 \texttt{\small travel}\rightarrow \texttt{\small Beijing} &\otimes& \texttt{\small Great-Wall} 
\end{eqnarray*}
which discriminates between word co-occurrence (e.g., \{\texttt{\small Beijing},\hspace{-3pt} \texttt{\small travel}\}) and dependency-tree pattern (e.g., $ \texttt{\small travel}\hspace{-3pt}\rightarrow \hspace{-3pt}\texttt{\small New York} $), and gives a higher score to (\textsf{\small T1}, \textsf{\small R1$_A$}).

The mining algorithm allows us to find patterns representing deep and long-distance relationship within two short-texts to be matched. The \emph{deep features} therefore provide sophisticated matching structures between two texts. In contrast, the \emph{shallow features} can only give word-level correspondences between words in two texts. The difference is analogous to syntax-based translation model and word-based translation model \cite{Koehn}.

\section{The Deep Matching Model} \label{s:matchmodel}
The dependency tree matching patterns (or deep features) are then incorporated into a deep neural network for determining the matching degree of a  pair of short-texts.

\subsection{Model Description} 
The diagram of our deep matching model is given in Fig.\ref{f:OverallDiagram}. When a pair of short-texts is given, we first obtain their dependency trees, form the direct product of them, and then perform abstraction on them (if suitable). After that, we look up the table of dependency tree matching patterns and convert the input text-pair into a binary vector, where an element is one if the corresponding pattern can apply to the input text-pair, otherwise it is zero. The binary vector, which is of $10M$-dimension and is sparse with typically 10$\sim$50 ones in our experiments, is then fed into the deep neural network for the final match decision. 
\subsection{Learning}
The learning of the deep neural network consists of 1) learning of the architecture, and 2) tuning of the parameters. 

\subsubsection{Architecture Learning} \label{s:A-Learning}

Since there are  $10M$  raw features, the number of parameters will be too large if we have the input layer fully connected to the first hidden layer with a reasonable size (say, 1,000 nodes).  It is therefore necessary to specify sensible sparse patterns to ensure that the information in the raw features can be well abstracted in the first hidden layer.

It is believed that neural networks are more suited for dense and continuous input, and there is little work on building an appropriate architecture for sparse and discrete input with a demanding size. In this work, we take a simple procedure to ensure each input node is connected to approximately $K$ (referred to as NodeDensity later in the paper) hidden nodes, and
the average activations of the hidden nodes (measured as the average times of them connected to hit features) are approximately the same. The underlying belief is that we can preserve as much information as possible when going from the sparse hit patterns to the dense 1,000-D representation. 

\paragraph{The Selection of Overall Architecture}
The overall architecture of the neural network is illustrated in Fig.\ref{f:OverallDiagram}. As it shows, we have 1,000 units in the first hidden layer (sigmoid active function), 400 in the second hidden layer, 30 in the third hidden layer, and one in the output layer.  Empirical results show that this architecture performs slightly better than a 3-layer one with approximately same number of parameters, while more hidden layers (say, 5) do not bring any significant further improvement.

\subsection{Parameter Learning}
We employ a discriminative training strategy with a large margin objective. Suppose that we are given the following triples $(\x, \y^+, \y^-)$ from the oracle, with $\x$ ($\in \calX$) matched with $\y^+$ better than with $\y^-$ (both $\in \calY$).  We have the following pairwise loss as objective:
{\small
\begin{equation*}
\calL(\calW, \calD_{trn}) = \sum_{(\x_i, \y_{i}^+, \y_{i}^-) \in \calD_{trn}}  \hspace{-5pt} e_{\calW}(\x_i, \y_{i}^+, \y_{i}^-) + R(\calW), 
\end{equation*}}
\noindent \hspace{-5pt} where $R(\calW)$ is the regularization term, and $e_{\calW}(\x_i, \y_{i}^+, \y_{i}^-)$ is the error for triple $(\x_i, \y_{i}^+, \y_{i}^-)$, given by the following large margin form:
{\small
\[
e_{\calW}(\x_i, \y_{i}^+, \y_{i}^-)\hspace{-1pt} = \hspace{-1pt}
\max(0, m+\textsf{s}(\x_i,\y_i^-)-\textsf{s}(\x_i,\y_i^+)), 
\]}

\noindent with $m$ ($0<m$) controlling the margin in training. In the experiments, we use $m=1$, but we find that the results are rather stable with $m$ in a fairly large range.

For training, we use the generic back-propagation algorithm adapted for the sparse patterns in the first layer. More specifically, when updating the weights in the first layer, we only update the weights associated with the active nodes in the input layer, which faithfully respects the law of back-prop but makes the learning efficiently enough even on a training set with millions of instances.  It is easy to see that the number of parameters (at least $4\times10^7$) is greater than the number of positive instances, and thus some kind of regularization is needed. Here we consider employing both dropout~\cite{dropout} and early stopping~ \cite{earlystoping}, which turns out to be important for the success of the model, especially when the number of parameters is over $10^8$.

\section{Experiments} \label{s:experiments} 
We report our empirical study of \textsc{DeepMatch}$_{tree}$ and compare it to competitors, with a brief analysis and case studies. 

\subsection{Datasets and Evaluation Metric} 
The experiments are on two Weibo datasets in two settings.  

\noindent \paragraph{Original-vs-Random:}
The first dataset, denoted as \textsf{\small DataOrignal}, consists of 4.8 million (tweet, response) pairs.
For each positive pair (original pair),  we randomly select ten responses as negative examples (contrastive sampling of negative examples), rendering 45 million triples. Our evaluation shows that for a given tweet there is $<$1\% chance that a randomly selected response out of 10 is suitable.

We use 485,282 original (tweet, response) pairs not used in the training for testing. For each pair, we get nine random responses, and testing the performance of each matching model to pick the correct response.  

\noindent \paragraph{Retrieval-based Conversation:}
The second dataset, denoted as \textsf{\small DataLabeled}, consists of 422 tweets and around 30 labeled responses for each tweet
\footnote{Data: \url{data.noahlab.com.hk/conversation/}}, as introduced in~\cite{emnlpmatch} for retrieval-based conversation.

On \textsf{\small DataLabeled}, we test how different matching models enhance the performance of the retrieval-based conversation model~\cite{emnlpmatch} on finding a suitable response for a given tweet. It is rather hard, since the negative responses are topically related to the tweet. We use the same retrieval strategy in~\cite{emnlpmatch}, while individually adding the scores of the matching models as a new feature of the ranking function to rank retrieved responses (20$\sim$30 for each tweet).

In both experiments we use precision at one (P@1) \cite{L2R} to measure the accuracy of matching. Basically, for each given tweet \textsf{\small T}, we calculate the matching scores between \textsf{\small T} and all candidate responses, and select the one with the highest score. The ranking gets one point iff the selected one is the original (on the \textsf{\small Original-vs-Random} dataset) or labeled as ``good" (on the \textsf{\small Retrieval-based Conversation} dataset). P@1 measures the chance of getting the selection right averaged over all the tweets in the test set. 

\subsection{Competitor Methods} 
\begin{itemize}
  \item \textsc{Translation:} We use the translation probability $p(\textsf{\small response}|\textsf{\small tweet})$ to measure the matching level between the response and tweet\footnote{This performs slightly better than $ p(\textsf{\scriptsize tweet}|\textsf{\scriptsize response})$.}, which is estimated on a variant of IBM model 1 \cite{ibmmodel} adapted for this task. 
  \item \textsc{CosSim:} We simply calculate the cosine similarity between two short-texts with their TF-IDF representations. This method is still better than random since a good response tends to share words with the original tweet; 
  \item \textsc{WordEmbed:} We represent each short-text as the sum of the embedding vectors of the words which it contains. The matching score of two short-texts is calculated using a multi-layer perceptron (MLP) with concatenation of the two vectors as input; 
  \item \textsc{DeepMatch$_{topic}$:} We employ the matching model in \cite{nipsmatch} on the basis of topics and train a neural network with 3 hidden layers and 1,000 hidden nodes in the first hidden layer;

  \item \textsc{DeepMatch$_{cnn}$:} We exploit the matching model proposed in \cite{NIPS2014_5550} represented as a convolutional neural network (CNN).
 
  \item \textsc{LR}$_{tree}$: To show the power of mined patterns we also train a logistic regression model taking all the mined patterns as input with the contrastive sampling training strategy. This can be viewed as a shallow version of \textsc{DeepMatch}$_{tree}$.

\end{itemize}
The methods can be roughly categorized into pattern-based methods (\textsc{CosSim}, \textsc{Translation}, \textsc{LR}$_{tree}$, \& \textsc{DeepMatch}$_{tree}$) and embedding-based methods (\textsc{WordEmbed}, \textsc{DeepMatch}, \& \textsc{DeepMatch}$_{cnn}$), where embedding-based methods represent each word with a vector, based on which the final matching decision is made.

All non-convex models are trained with stochastic gradient descent (SGD) \cite{sgd}.We find that their performances are in general quite insensitive to the size of mini-batch.

\subsection{Results on Original-vs-Random}
In this section we present the results in the orig nal-vs-random setting.
For each model, we only report its best performance on the test data, since the large size of test data removes any chance of ``accidental cheating". We first study the architecture variations of \textsc{DeepMatch}$_{tree}$, and then compare its best setting against the compe放titors.

Here we compare the performances of \textsc{DeepMatch}$_{tree}$ under different settings, more specially, the number of hidden layers (1$\sim$5), NodeDensity (1$\sim$20) and architecture learning (details of results are omitted).
In a nutshell, the performance peaks around NodeDensity=10 with architecture learning.  With NodeDensity$\geq 10$, the matching model has over $10^8$ parameters, and it needs regularization (e.g., dropout) to prevent overfitting in addition to early stopping. The influence of architecture learning is most salient for a relatively large NodeDensity (say, $>3$), while the number of hidden layers stops bringing significant improvement when $\geq 3$. Generally we found that architectures deeper and larger than the current one does not bring any significant improvement but much slower. 
%

\subsubsection{Comparison to Competitor Models}
Table \ref{t:ResultsAllModels} compares \textsc{DeepMatch}$_{tree}$ to the competitor models. As it shows, our model outperforms all the competitor models with large margins. The contribution of  deep architectures is manifested by the differences between the deep architectures and shallow ones with the same mined patterns.
\begin{table}[h!]
  \centering
\begin{tabular}{l|l|l}
  \hline
  Model & P@1 (1v1) & P@1 (1v9)\\
  \hline
  \textsc{CosSim} & 0.554 & 0.377  \\
  \hline
  \textsc{DeepMatch}$_{topic}$ & 0.701 & 0.330  \\
  \hline
  \textsc{WordEmbed} & 0.774 & 0.370  \\
  \hline
  \textsc{Translation} & 0.819 & 0.586  \\
  \hline
  \textsc{DeepMatch$_{cnn}$} & 0.851 & 0.496 \\
  \hline \hline
  \textsc{LR}$_{tree}$ & 0.853 & 0.652  \\
  \hline
  \textsc{DeepMatch}$_{tree}$ & {\bf 0.889} & {\bf 0.708}  \\
  \hline
\end{tabular}
   \caption{The results of all models on Original-vs-Random. {DeepMatch}$_{tree}$ signiﬁcantly  outperforms all the baselines ($p < 0.01$ from t-test).
   }
   \label{t:ResultsAllModels} 
\end{table}

There is a vast gap between pattern-based models and embedding-based models. Although the embedding-based methods can perform fairly well  on the one versus one (1v1) setting (0.85+), the performance drops dramatically in the one versus nine (1v9) settings (0.49+), while the pattern-based methods can maintain over 0.55 (dropping from 0.80+) in the same test setting. This contrast suggests that pattern-based methods, with varying coverage in the feature space, are more certain on ``matched" positive cases than on negative cases, yielding more reliable ranking results.

\subsection{Results on Conversation Data} 
For each model, we use 5-fold cross validation to choose the hyper-parameter of the ranking model RankSVM and report the best result. Clearly \textsc{DeepMatch}$_{tree}$ can greatly improve the performances of retrieving a suitable response from the pool, with  significantly better accuracies over the competitor models($p < 0.05$ from t-test). This result is consistent with the result on Original-vs-Random despite the difference in experimental setting.
\begin{table}[h!]
  \centering
\begin{tabular}{l|l}
  \hline
  Model & P@1   \\ \hline
  \textsc{Baseline} & 0.574  \\
  \hline
  +\textsc{DeepMatch}$_{topic}$ & 0.587  \\
  \hline
  +\textsc{WordEmbed} & 0.579  \\
  \hline
  +\textsc{Translation} & 0.585  \\
  \hline
  +\textsc{DeepMatch}$_{tree}$ & {\bf 0.608} \\
  \hline
\end{tabular}
   \caption{The results on retrieval-based conversation.}
   \label{f:conversation} 
\end{table}

\subsection{Analysis and Case Study} 
\noindent \paragraph{Deep vs. Shallow Patterns}
Deep patterns represent information that cannot be adequately modeled by shallow patterns in a deep neural network. Indeed, our study shows that on Original-vs-Random data, P@1 decreases to 0.871 (1v1) and 0.688 (1v9) after removing the deep features. Below is a real case in our experiment.  This observation is interesting since feature learning is previously often taken as partially the responsibility of deep learning.
\begin{figure}[h!]
\begin{center}
    \begin{tabular}[l]{cc}
    \includegraphics[width=0.38\textwidth]{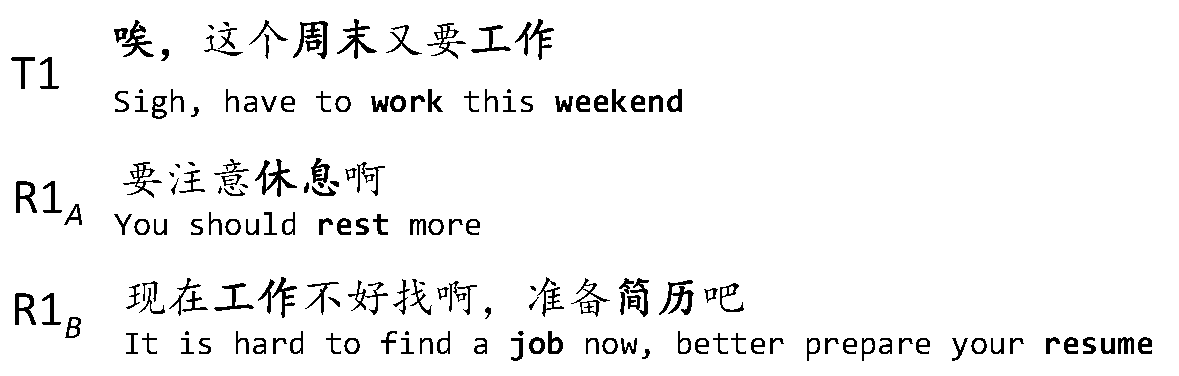}
     \end{tabular} 
  \end{center}
\end{figure} 

\noindent When trying to find a matched response  for \textsf{\small T1} , the ``deep" pattern \{$\texttt{\small work} \hspace{-3pt} \rightarrow \hspace{-3pt}\texttt{\small weekend} \; \otimes \; \texttt{\small rest} $\} plays a determining role in picking \textsf{\small R1$_A$} over \textsf{\small R1$_B$}, while shallower patterns as  \{$\texttt{\small work} \; \otimes \; \texttt{\small job} $\} and  \{$\texttt{\small work} \; \otimes \; \texttt{\small resume} $\} favor \textsf{\small R1$_B$}.

\noindent \paragraph{The effect of abstraction}
The abstraction step helps improve the generalization ability of the matching model, by improving P@1 on Original-vs-Random from 0.876 to 0.889 (1v1)  and 0.694 to 0.708 (1v9). This can also be illustrated with the following real example from our experiment, 
\begin{figure}[h!]
\begin{center}
    \begin{tabular}[l]{cc}
    \includegraphics[width=0.38\textwidth]{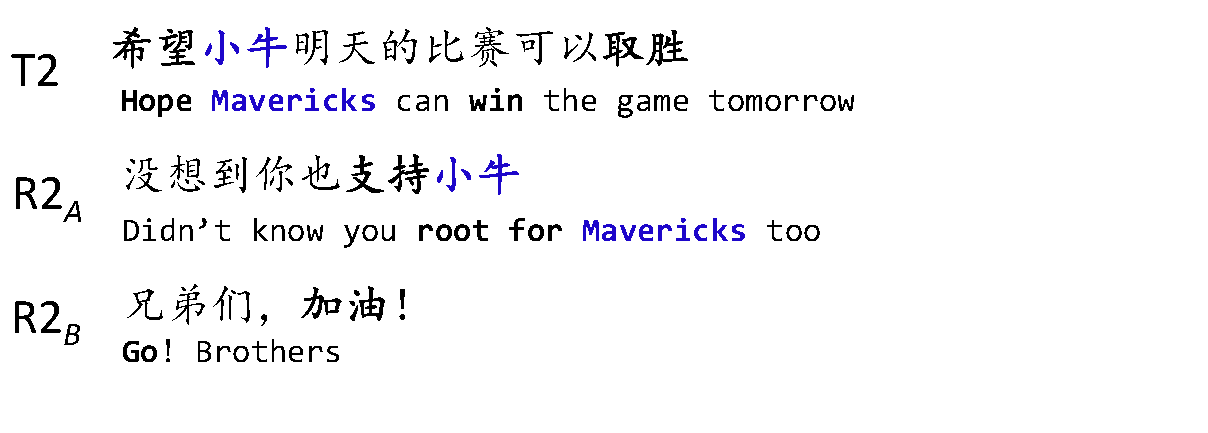}
     \end{tabular} 
  \end{center} 
\end{figure}

\noindent Suppose that for \textsf{\small T2} we want to pick a more appropriate response from candidates \textsf{\small R2$_A$} and \textsf{\small R2$_B$}. The mining algorithm (Algorithm 1) discovers the following pattern after the SameEntity abstraction 
\[
\texttt{\small hope} \rightarrow \texttt{\small win} \rightarrow x \; \otimes \; \texttt{\small support} \rightarrow x 
\]
where $x$ stands for any named entity. This pattern (and its own sub-patterns) then plays an important role in the later  matching model in assigning a higher matching score to (\textsf{\small T2}, \textsf{\small R2$_A$}), covering  more specific patterns like 
\[
\texttt{\small hope}\hspace{-3pt} \rightarrow \hspace{-3pt}\texttt{\small win} \hspace{-3pt}\rightarrow \hspace{-3pt}\texttt{\small Mavericks} \; \otimes \; \texttt{\small support} \hspace{-3pt}\rightarrow \hspace{-3pt}\texttt{\small Mavericks} 
\]
which are filtered out in the mining step for its small support. 


\section{Related Work} \label{s:related} 
The proposed model is related to several threads of work in natural language processing and machine learning. 
\paragraph{Deep Matching Models}
There are other works on using deep neural networks for the matching task~\cite{CIKM,bordes2014semantic,Sun_2013_ICCV,nipsmatch,NIPS2014_5550}, which build upon given or learned representations of objects. In our model, we try to directly mine and learn the representations of matching. 
\paragraph{Graph-based Kernel}
\textsc{DeepMatch}$_{tree}$ extends the important notions in conventional graph kernels \cite{VishproG} in two senses. First, our model allows matching of two different subgraphs in two domains (e.g., \{\texttt{\small work}$\rightarrow$ \texttt{\small weekend}\} in one domain and  \{\texttt{\small have}$\rightarrow$ \texttt{\small rest}\} in the other), while graph kernels only consider the common subgraphs on two sides. Second, our model captures the nonlinear and hierarchical relations between different matching patterns, while graph kernels simply add them together, with different weights determined by the types of sub-graphs.
\paragraph{String-Rewriting Kernel}
\textsc{DeepMatch}$_{tree}$  is also related to the string-rewriting kernel (SRK) \cite{SRK} for paraphrase identification, in that SRK also generates many patterns of matching and learns to weigh them in training. The main difference is the matching patterns considered in SRK are exhaustively enumerated (although calculated in a smart way), while ours are discovered via a mining algorithm. 

\section{Conclusion} 
We propose a generic model for matching two short-texts, which relies on a tree-mining algorithm to discover a vast amount of matching patterns and a DNN to further perform the task using those patterns. Empirical study on the rather difficult task of tweet and response matching shows that our model can outperform competitor with large margins.
\section{Acknowledge}
This work is supported in part by China National 973 project 2014CB340301. Qun Liu's work is partially supported by the Science Foundation 
Ireland (Grant 12/CE/I2267 and 13/RC/2106) as part of the ADAPT Centre 
at Dublin City University.
\bibliographystyle{named}
\small{\bibliography{ijcai}}
\end{document}